\documentclass[letterpaper, 10 pt, conference]{ieeeconf}  

\IEEEoverridecommandlockouts                              

\usepackage{graphicx}
\usepackage{caption}      
\usepackage{subcaption}   

\DeclareCaptionLabelFormat{tablel}{TABLE #2}
\renewcommand{\thesubtable}{\Roman{subtable}}
\makeatletter
\renewcommand\p@subtable{\thesubtable~}
\makeatother

\usepackage{algorithm}
\usepackage{listings}
\usepackage[noend]{algpseudocode}
\usepackage{wrapfig}
\usepackage{multirow}
\usepackage{hyperref}
\usepackage{tabularx}

\title{\LARGE \bf
Multi-Modal Geometric Learning for Grasping and Manipulation
}

\author{David Watkins-Valls, Jacob Varley, and Peter Allen
\thanks{This work is supported by NSF Grant CMMI 1734557. We gratefully acknowledge the support of NVIDIA Corporation with the donation of the Titan Xp GPU used for this research. Authors are with Columbia University, 
        {\tt\small (davidwatkins, jvarley, allen)@cs.columbia.edu}}
}

\begin{document}

\maketitle
\thispagestyle{empty}
\pagestyle{empty}

\begin{abstract}

This work provides an architecture that incorporates depth and tactile information to create rich and accurate 3D models useful for robotic manipulation tasks. This is accomplished through the use of a 3D convolutional neural network (CNN).  Offline, the network is provided with both depth and tactile information and trained to predict the object's geometry, thus filling in regions of occlusion.  At runtime, the network is provided a partial view of an object. Tactile information is acquired to augment the captured depth information. The network can then reason about the object's geometry by utilizing both the collected tactile and depth information. We demonstrate that even small amounts of additional tactile information can be incredibly helpful in reasoning about object geometry. This is particularly true when information from depth alone fails to produce an accurate geometric prediction.  Our method is benchmarked against and outperforms other visual-tactile approaches to general geometric reasoning.  We also provide experimental results comparing grasping success with our method.

\end{abstract}

\section{Introduction}

Robotic grasp planning based on raw sensory data is difficult due to occlusion and incomplete information regarding scene geometry.  Often, for example, one sensory modality does not provide enough context to enable reliable planning.  For example, a single depth sensor image cannot provide information about occluded regions of an object, and tactile information is incredibly sparse.  This work utilizes a 3D convolutional neural network to enable stable robotic grasp planning by incorporating both tactile and depth information to infer occluded geometries.  This multi-modal system is able to utilize both tactile and depth information to form a more complete model of the space the robot can interact with and also to provide a complete object model for grasp planning.

At runtime, a point cloud of the visible portion of the object is captured, and multiple guarded moves are executed in which the hand is moved towards the object, stopping when contact with the object occurs. The newly acquired tactile information is combined with the original partial view, voxelized, and sent through the CNN to create a hypothesis of the object's geometry.

\begin{figure}[t]
    \centering
    \includegraphics[width=.48\textwidth] {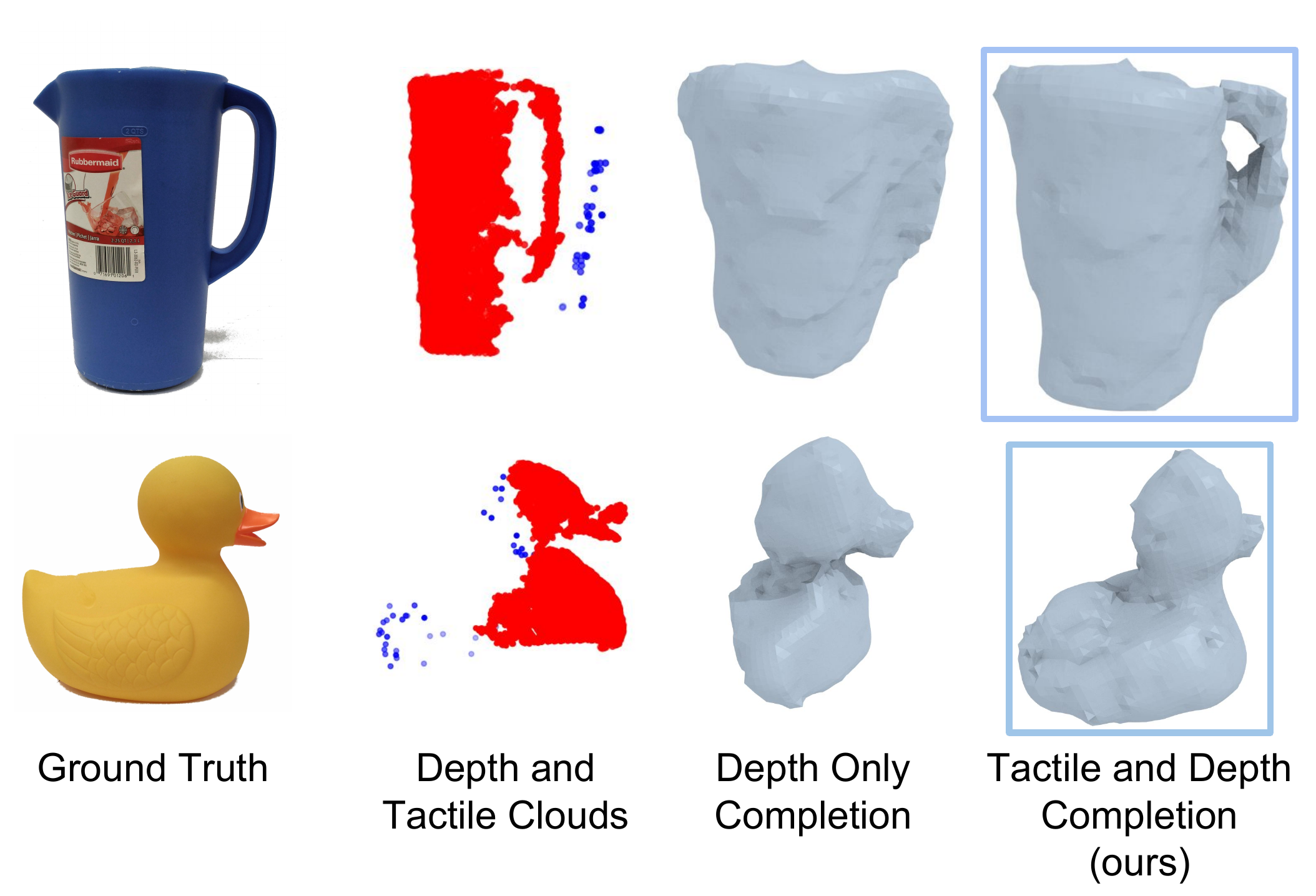}
    \caption{Completion example from tactile and depth data. These completions demonstrate that small amounts of additional tactile sensory data can significantly improve the system's ability to reason about 3D geometry. The Depth Only Completion for the pitcher does not capture the handle well, whereas the tactile information gives a better geometric understanding. For this example, the additional tactile information allowed the CNN to correctly identify a handle in the completion mesh and similar completion improvement was found for the rubber duck. The rubber duck was not present in the training data.  }
    \label{fig:cover_pic}
\end{figure}

\begin{figure*}[t]
\vspace{2mm}

    \centering
    \includegraphics[width=.95\textwidth] {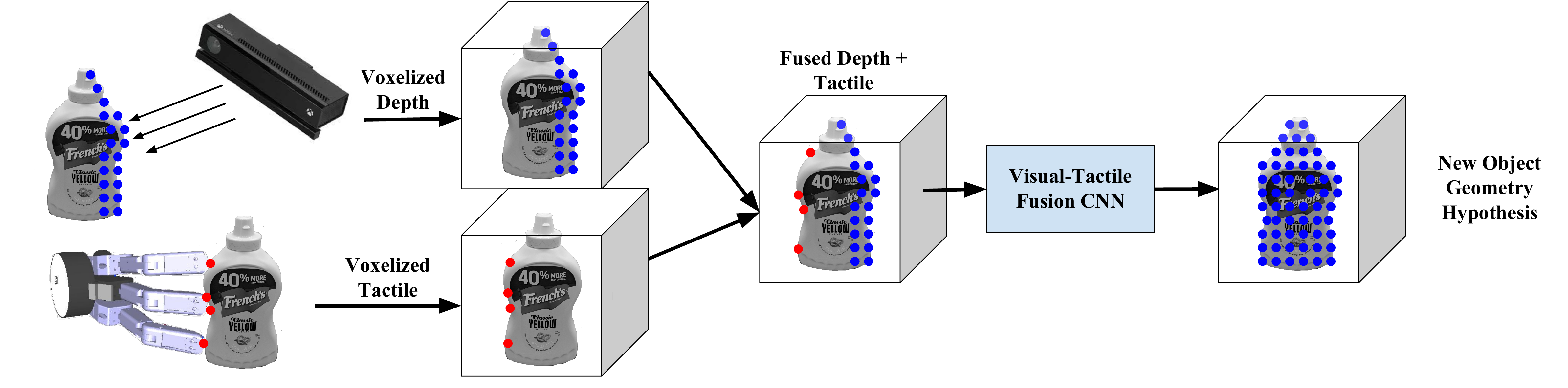}
    \caption{Both tactile and depth information are independently captured and voxelized into $40^3$ grids.  These are merged into a shared occupancy map which is fed into a CNN to produce a hypothesis of the object's geometry.}
    \label{fig:visual_tactile_fusion_cnn}
    \vspace{-4mm}
\end{figure*}

Depth information from a single point of view often does not provide enough information to accurately predict object geometry. There is often unresolved uncertainty about the geometry of the occluded regions of the object. To alleviate this uncertainty, we utilize tactile information to generate a new, more accurate hypothesis of the object's 3D geometry, incorporating both visual and tactile information. Fig. \ref{fig:cover_pic} demonstrates an example where the understanding of the object's 3D geometry is significantly improved by the additional sparse tactile data collected via our framework. An overview of our sensory fusion architecture is shown in Fig. \ref{fig:visual_tactile_fusion_cnn}.

This work is differentiated from others \cite{wang2018gelsighttactile} in that our CNN is acting on both the depth and tactile as input information fed directly into the model rather than using the tactile information to update the output of a CNN not explicitly trained on tactile information. This enables the tactile information to produce non-local changes in the resulting mesh. In many cases, depth information alone is insufficient to differentiate between two potential completions, for example a pitcher vs a rubber duckie. In these cases, the CNN utilizes sparse tactile information to affect the entire completion, not just the regions in close proximity to the tactile glance. If the tactile sensor senses the occluded portion of a drill, the CNN can turn the entire completion into a drill, not just the local portion of the drill that was touched. 

The contributions of this work include: 1) a framework for integrating multi-modal sensory data to holistically reason about object geometry and enable robotic grasping, 2) an open source dataset for training a shape completion system using both tactile and depth sensory information, 3) open source code for alternative visual-tactile general completion methods, 4) experimental results comparing the completed object models using depth only, the combined depth-tactile information, and various other visual-tactile completion methods, and 5) real and simulated grasping experiments using the completed models. This  dataset, code, and extended video  are  freely  available  at \url{http://crlab.cs.columbia.edu/visualtactilegrasping/}. 

\section{Related Work}

The idea of incorporating sensory information from vision, tactile and force sensors is not new \cite{miller1999integration}.  Despite the intuitiveness of using multi-modal data, there is still no concensus on which framework best integrates multi-modal sensory information in a way that is useful for robotic manipulation tasks. While prior work has been done to complete geometry using depth alone, none of these works consider tactile information\cite{dai2017complete}\cite{Dharmasiri2018}. In this work, we are interested in reasoning about object geometry, and in particular, creating models from multi-modal sensory data that can be used for grasping and manipulation.

Several recent uses of tactile information to improve estimates of object geometry have focused on the use of Gaussian Process Implicit Surfaces (GPIS) \cite{williams2007gaussian}.  Several examples along this line of work include \cite{caccamo2016active}\cite{yi2016active} \cite{bjorkman2013enhancing}\cite{dragiev2011gaussian}\cite{jamali2016active}\cite{sommer2014bimanual}\cite{mahler2015gp}.  This approach is able to quickly incorporate additional tactile information and improve the estimate of the object's geometry local to the tactile contact or observed sensor readings.  There has additionally been several works that incorporate tactile information to better fit planes of symmetry and superquadrics to observed point clouds \cite{ilonen2014three}\cite{ilonen2013fusing}\cite{bierbaum2008robust}.  These approaches work well when interacting with objects that conform to the heuristic of having clear detectable planes of symmetry or are easily modeled as superquadrics. 

There has been successful research in utilizing continuous streams of visual information similar to Kinect Fusion \cite{newcombe2011kinectfusion} or SLAM \cite{thrun2008simultaneous} in order to improve models of 3D objects for manipulation, an example being \cite{krainin2011manipulator}\cite{krainin2011autonomous}. In these works, the authors develop an approach to building 3D models of unknown objects based on a depth camera observing the robot's hand while moving an object. The approach integrates both shape and appearance information into an articulated ICP approach to track the robot's manipulator and the object while improving the 3D model of the object.  Similarly, another work \cite{hermann2016eye} attaches a depth sensor to a robotic hand and plans grasps directly in the sensed voxel grid. These approaches improve their models of the object using only a single sensory modality but from multiple points in time. 

In previous work \cite{varley2017shapecompletion_iros}, we created a shape completion method using single depth images. The work provides an architecture to enable robotic grasp planning via shape completion, which was accomplished through the use of a 3D CNN. The network was trained on an open source dataset of over 440,000 3D exemplars captured from varying viewpoints. At runtime, a 2.5D point cloud captured from a single point of view was fed into the CNN, which fills in the occluded regions of the scene, allowing grasps to be planned and executed on the completed object. The runtime of shape completion is rapid because most of the computational costs of shape completion are borne during offline training. This prior work explored how the quality of completions vary based on several factors. These include whether or not the object being completed existed in the training data, how many object models were used to train the network, and the ability of the network to generalize to novel objects, allowing the system to complete previously unseen objects at runtime. The completions are still limited by the training datasets and occluded views that give no clue to the unseen portions of the object.  From a human perspective, this problem is often alleviated by using the sense of touch.  In this spirit, this paper addresses this issue by incorporating sparse tactile data to better complete the object models for grasping tasks.

\section{Visual-Tactile Geometric Reasoning}
Our framework utilizes a trained CNN to produce a mesh of the target object, incorporating both depth and tactile information. 
We utilize the same architecture as found in \cite{varley2017shapecompletion_iros}. The model was implemented using the Keras \cite{Keras2015} deep learning library. Each layer used rectified linear units as nonlinearities except the final fully connected (output) layer which used a sigmoid activation to restrict the output to the range $[0,1]$. We used the cross-entropy error $E(y,y^\prime)$ as the cost function with target $y$ and output $y^\prime$:
$$E(y,y^\prime)=-\left( y \log(y^\prime) + (1 - y) \log(1 - y^\prime) \right) $$

\begin{figure}[t]
\vspace{2mm}
\centering

	\begin{subfigure}{.23\textwidth}
		\centering
		\includegraphics[width=1\textwidth]{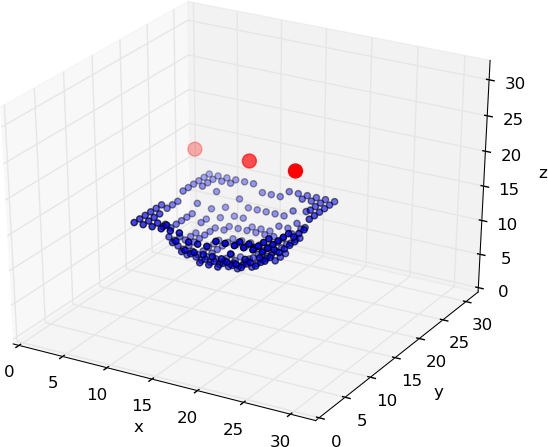}
	\end{subfigure}
	\begin{subfigure}{.23\textwidth}
		\centering
		\includegraphics[width=1\textwidth]{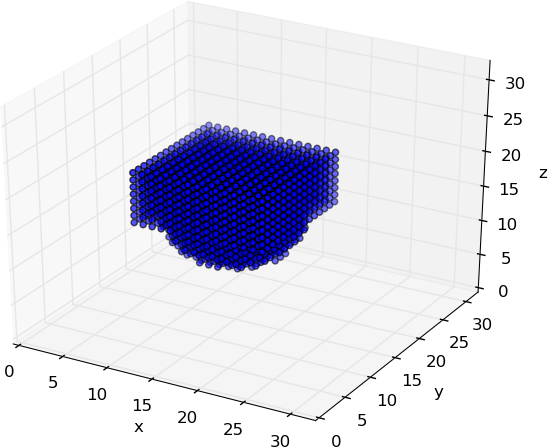}
	\end{subfigure}
	\caption{Example training pair from the geometric shape dataset. For the left, red dots represent tactile readings and blue dots represent the the depth image. The blue points on the right are the ground truth 3D geometry.} 
	\label{fig:geo_shape_training_example} 
	
\end{figure}

\begin{figure}[t]
\begin{algorithm}[H]
\caption{Simulated YCB/Grasp Tactile Data Generation}
\label{alg:TactileDataGenerationUniform}
\begin{algorithmic}[1]
\Procedure{Sample\_Tactile}{vox\_gt} \label{line:alg1line1}
\State grid\_dim = 40 // resolution of voxel grid \label{line:alg1line2}
\State npts = 40 // num locations to check for contact \label{line:alg1line3}
\State vox\_gt\_cf = align\_gt\_to\_depth\_frame(vox\_gt) \label{line:alg1line4}
\State xs = rand\_ints(start=0, end=grid\_dim-1, size=npts) \label{line:alg1line5}
\State ys = rand\_ints(start=0, end=grid\_dim-1, size=npts) \label{line:alg1line6}
\State tactile\_vox = [] \label{line:alg1line7}
\For{x, y in xs, ys} \label{line:alg1line8}
    \For{z in range(grid\_dim-1, -1, -1)} \label{line:alg1line9}
        \If{vox\_gt\_cf[x, y, z] == 1} \label{line:alg1line10}
        \State tactile\_vox.append(x, y, z) \label{line:alg1line11}
        \State continue \label{line:alg1line12}
        \EndIf
    \EndFor
\EndFor
\State tactile\_points = vox2point cloud(tactile\_vox) \label{line:alg1line13}
\State \Return tactile\_points \label{line:alg1line14}
\EndProcedure
\end{algorithmic}
\end{algorithm}
\vspace{-8mm}
\end{figure}

\noindent This cost function encourages each output to be close to either 0 for unoccupied target voxels or 1 for occupied target voxels. The optimization algorithm Adam \cite{kingma2014}, which computes adaptive learning rates for each network parameter, was used with default hyperparameters ($\beta_1$$=$$0.9$, $\beta_2$$=$$0.999$, $\epsilon$$=$$10^{-8}$) except for the learning rate, which was set to 0.0001. Weights were initialized following the recommendations of \cite{he2015} for rectified linear units and \cite{glorot2010} for the logistic activation layer. The model was trained with a batch size of 32. We used the Jaccard similarity \cite{jaccard} to evaluate the similarity between a generated voxel occupancy grid and the ground truth.


\section{Completion of Simulated Geometric Shapes}
\label{sec:geometric_shapes}



\begin{figure*}[t]
\vspace{2mm}
	\centering
	\begin{subfigure}[t]{0.3\textwidth}
		\centering
		\includegraphics[width=1\textwidth]{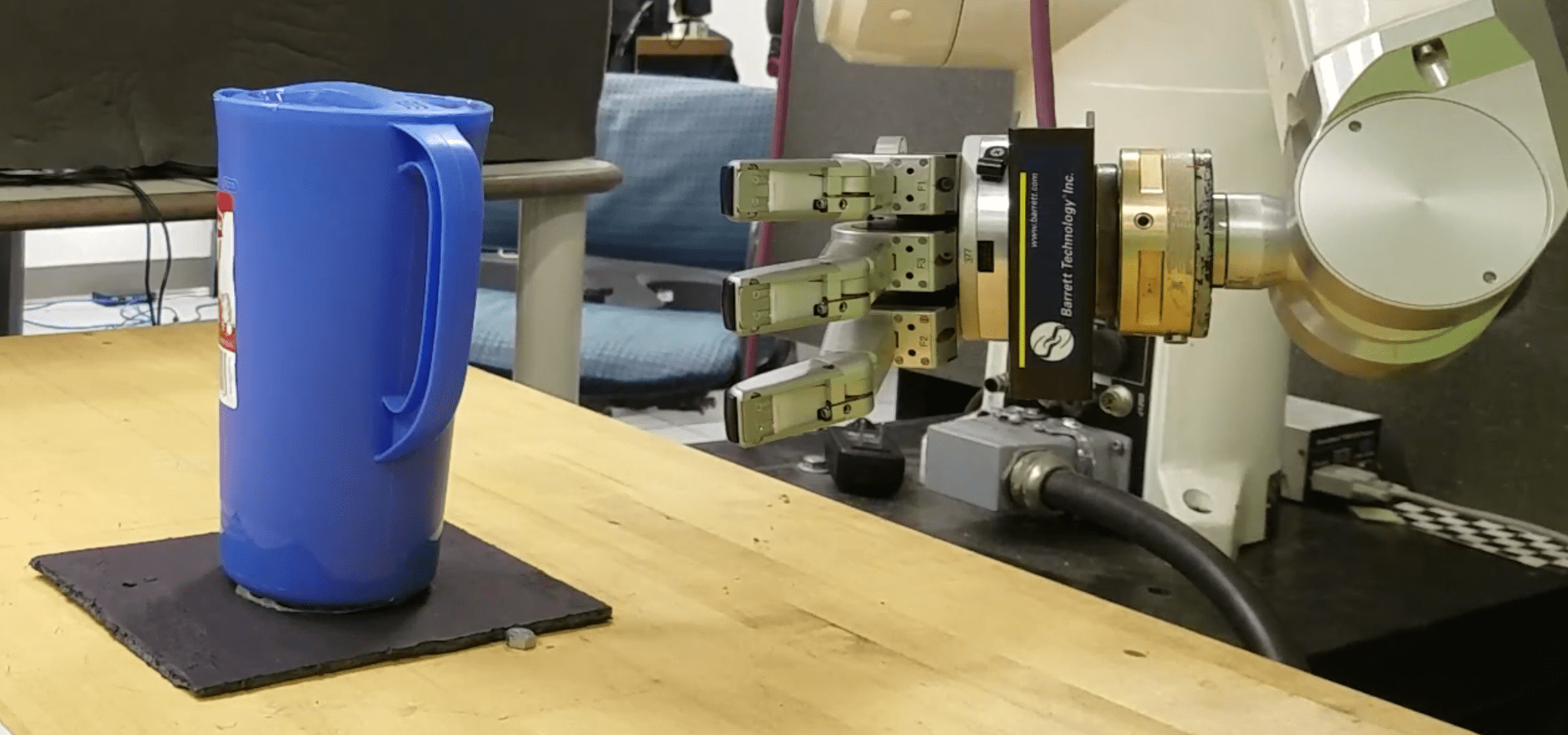}
		\caption{Hand approach}
	\end{subfigure}
	\begin{subfigure}[t]{0.3\textwidth}
		\centering
		\includegraphics[width=1\textwidth]{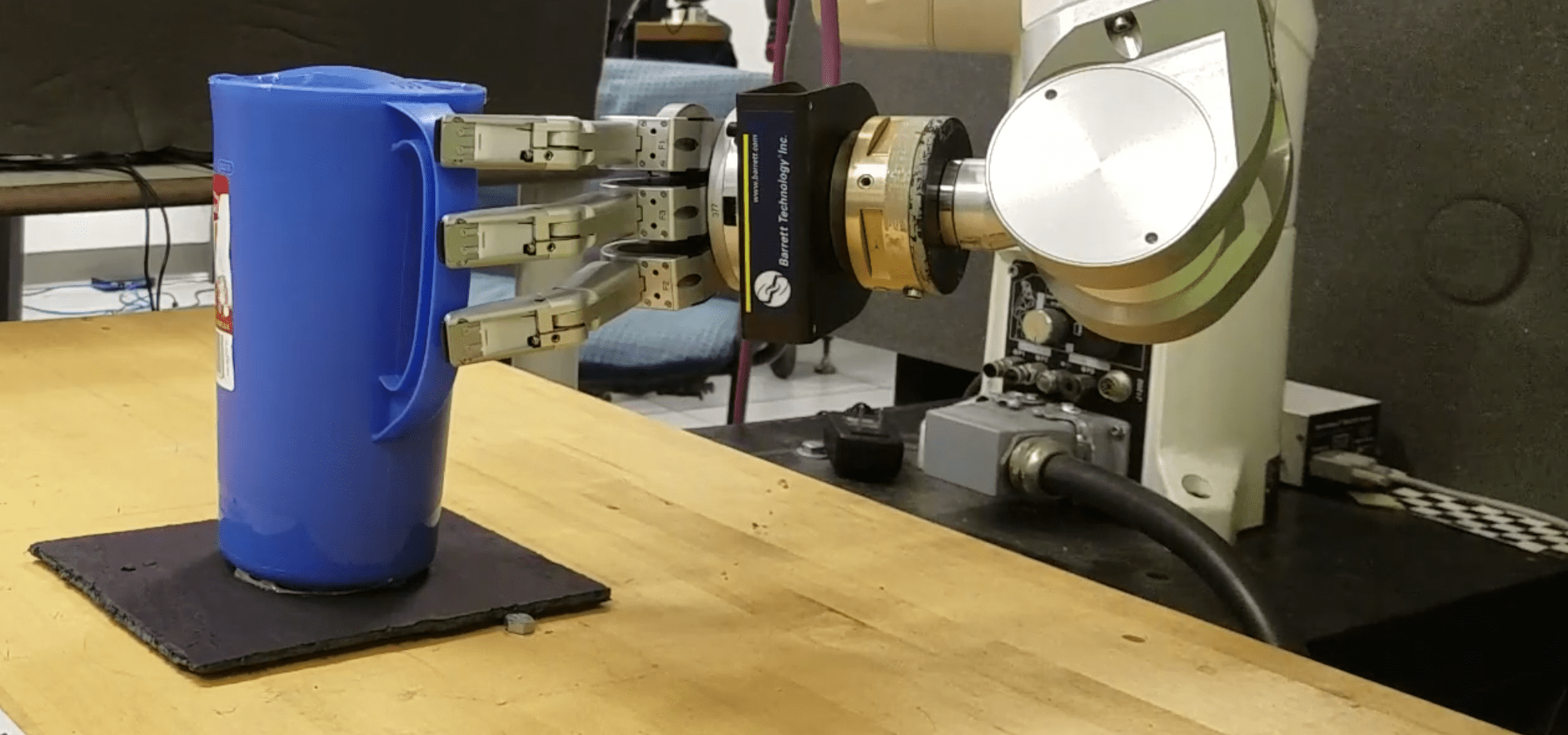}
		\caption{Finger contact}
	\end{subfigure}
	\begin{subfigure}[t]{0.3\textwidth}
		\centering
		\includegraphics[width=1\textwidth]{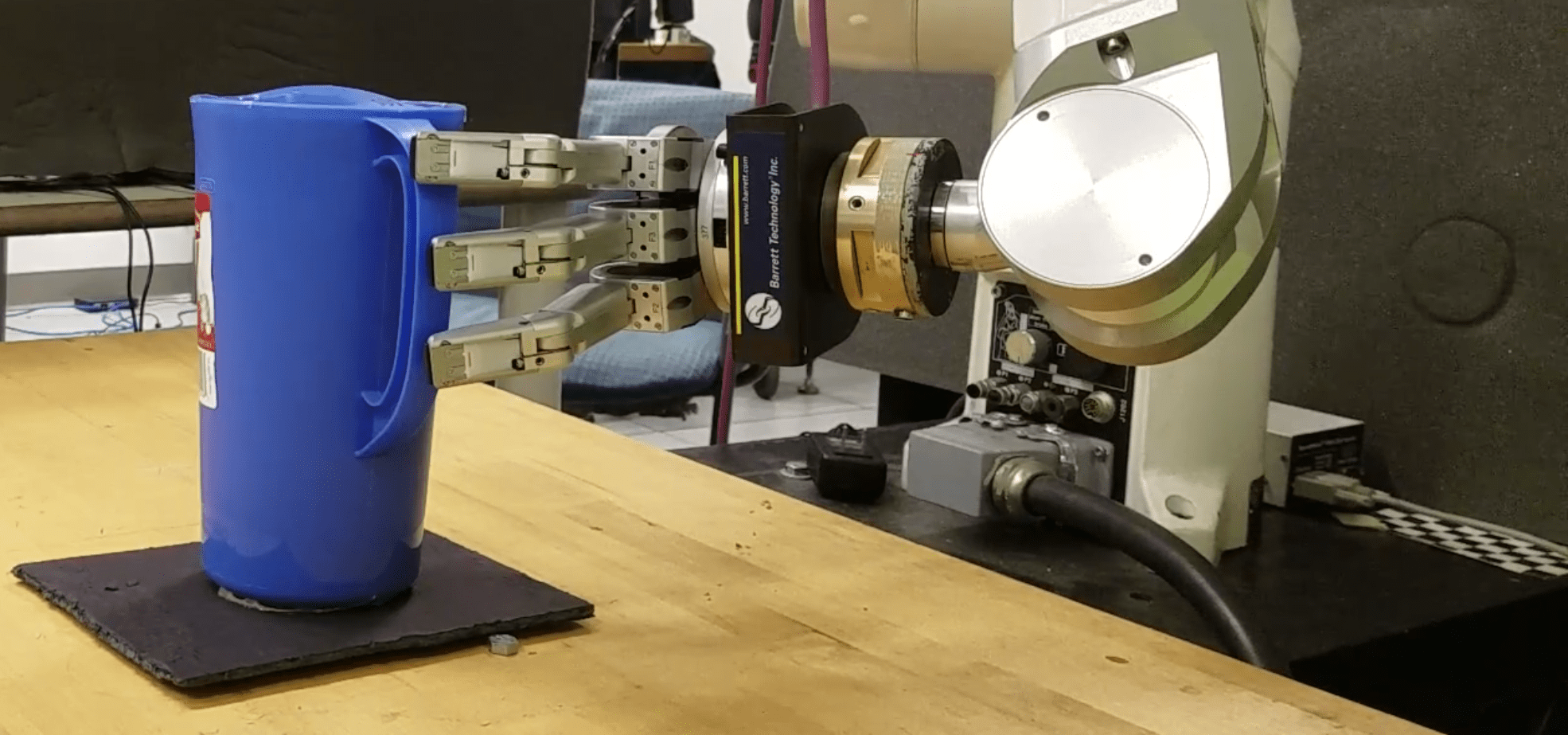}
		\caption{Finger curl}
	\end{subfigure}
	\caption{Barrett hand showing contact with a fixed object. (a) The hand is manually brought to an approach position, (b) approaches the object, and (c) the fingers are curled to contact the object and collect tactile information. This process is repeated 6 times over the occluded surface of the object. } 
	\label{fig:handobjectcontact} 
	\vspace{-4mm}
\end{figure*}

Three networks with the architecture from \cite{varley2017shapecompletion_iros} were trained on a simulated dataset of geometric shapes (Fig. \ref{fig:geo_shape_training_example}) where the front and back were composed of two differing shapes. Sparse tactile data was generated by randomly sampling voxels along the occluded side of the voxel grid. We trained a network that only utilized tactile information. This performed poorly due to the sparsity of information. A second network was given only the depth information during training and performed better than the tactile-only network did. It still encountered many situations where it did not have enough information to accurately complete the obstructed half of the object.  A third network was given depth and tactile information which successfully utilized the tactile information to differentiate between plausible geometries of occluded regions. 

The Jaccard similarity improved from 0.890 in the depth only network to 0.986 in the depth and tactile network.  This task demonstrated that a CNN can be trained to leverage sparse tactile information to decide between multiple object geometry hypotheses. When the object geometry had sharp edges in its occluded region, the system would use tactile information to generate a completion that contained similar sharp edges in the occluded region. This completion is more accurate not just in the observed region of the object but also in the unobserved portion of the object.

\begin{figure}[hbtp]
\centering
    \includegraphics[width=0.48\textwidth]{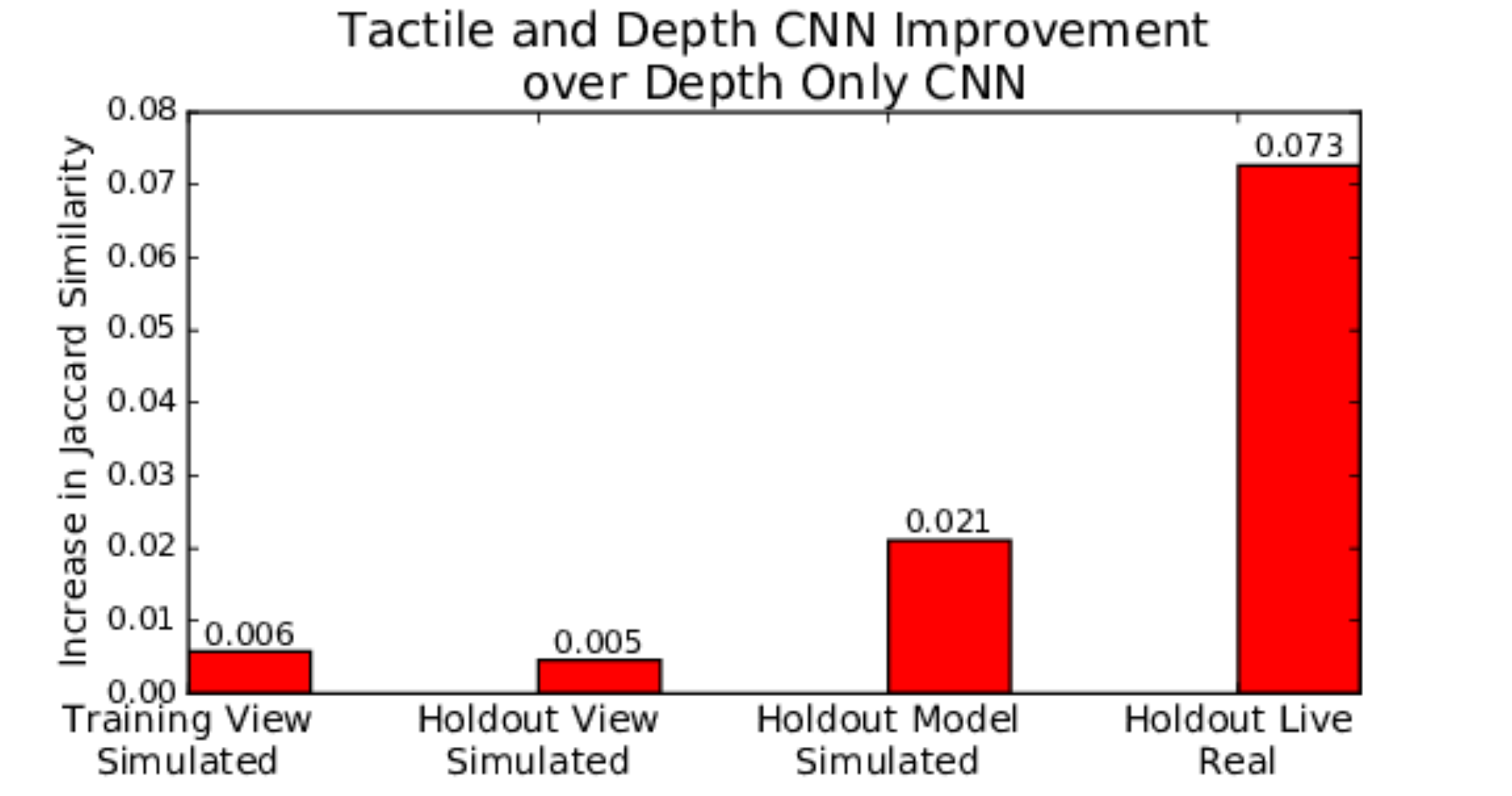}
    \caption{ As the difficulty of the data splits increase, the delta between the \textbf{Depth Only} CNN completion accuracy and the \textbf{Tactile and Depth} CNN completion accuracy increases.  The additional tactile information is more useful on more difficult completion problems. }
    \label{fig:jaccard_improvement}
    \vspace{-6mm}
\end{figure}

\setcounter{subtable}{0}
\begin{table*}[!htb]
\vspace{2mm}

\begin{tabularx}{\textwidth}{cc}
\begin{subtable}[t]{.48\linewidth}
    \begin{tabular}{|c|c|c|c|c|}
    \hline
    \multicolumn{1}{|c|}{\begin{tabular}[c]{@{}c@{}}\textbf{Completion} \\  \textbf{Method}\end{tabular}} 
    & \multicolumn{1}{c|}{\begin{tabular}[c]{@{}c@{}}\textbf{Train} \\  \textbf{View(Sim)}\end{tabular}} 
    & \multicolumn{1}{c|}{\begin{tabular}[c]{@{}c@{}}\textbf{Holdout} \\  \textbf{View(Sim)}\end{tabular}} 
    & \multicolumn{1}{c|}{\begin{tabular}[c]{@{}c@{}}\textbf{Holdout} \\  \textbf{Model(Sim)}\end{tabular}} 
    & \multicolumn{1}{c|}{\begin{tabular}[c]{@{}c@{}}\textbf{Holdout} \\  \textbf{(Live)}\end{tabular}} \\ 
    \hline
    	Partial     & 0.01          & 0.02          & 0.01          & 0.01          \\ \hline
    	Convex Hull & 0.50          & 0.51          & 0.46          & 0.43          \\ \hline
    	GPIS        & 0.47          & 0.45          & 0.35          & 0.48          \\ \hline
    	Depth CNN   & 0.68          & 0.65          & 0.65          & 0.37          \\ \hline
    	Ours        & \textbf{0.69} & \textbf{0.66} & \textbf{0.65} & \textbf{0.64} \\ \hline
    \end{tabular}
    
    \caption{\textbf{Jaccard similarity results}, measuring the intersection over union of two voxelized meshes, as described in Section \ref{sec:Completion_results}. (Larger is better)}
    \label{tab:Jaccard}
\end{subtable}&

\begin{subtable}[t]{.48\linewidth}
    \begin{tabular}{|c|c|c|c|c|c|c|}
    \hline
    \multicolumn{1}{|c|}{\begin{tabular}[c]{@{}c@{}}\textbf{Completion} \\  \textbf{Method}\end{tabular}} 
    & \multicolumn{1}{c|}{\begin{tabular}[c]{@{}c@{}}\textbf{Train} \\  \textbf{View(Sim)}\end{tabular}} 
    & \multicolumn{1}{c|}{\begin{tabular}[c]{@{}c@{}}\textbf{Holdout} \\  \textbf{View(Sim)}\end{tabular}} 
    & \multicolumn{1}{c|}{\begin{tabular}[c]{@{}c@{}}\textbf{Holdout} \\  \textbf{Model(Sim)}\end{tabular}} 
    & \multicolumn{1}{c|}{\begin{tabular}[c]{@{}c@{}}\textbf{Holdout} \\  \textbf{(Live)}\end{tabular}} \\ 
    \hline
    	Partial     & 7.8           & 7.0           & 7.6           & 11.9          \\ \hline
    	Convex Hull & 32.7          & 45.1          & 49.1          & 11.6          \\ \hline
    	GPIS        & 59.9          & 79.2          & 118.0         & 17.9          \\ \hline
    	Depth CNN   & 6.5           & 6.9           & 6.5           & 16.5          \\ \hline
    	Ours        & \textbf{5.8}  & \textbf{5.8}  & \textbf{6.2}  & \textbf{7.4}  \\ \hline
    \end{tabular}
    
    \caption{\textbf{Hausdorff distance results}, measuring the mean distance in millimeters from points on one mesh to points on another mesh, as described in Section \ref{sec:Completion_results}.  (Smaller is better)}
    \label{tab:Hausdorff} 
\end{subtable}\\

\begin{subtable}[t]{.48\linewidth}

    \begin{tabular}{|c|c|c|c|c|c|c|}
    \hline
    \multicolumn{1}{|c|}{\begin{tabular}[c]{@{}c@{}}\textbf{Completion} \\  \textbf{Method}\end{tabular}} 
    & \multicolumn{1}{c|}{\begin{tabular}[c]{@{}c@{}}\textbf{Train} \\  \textbf{View(Sim)}\end{tabular}} 
    & \multicolumn{1}{c|}{\begin{tabular}[c]{@{}c@{}}\textbf{Holdout} \\  \textbf{View(Sim)}\end{tabular}} 
    & \multicolumn{1}{c|}{\begin{tabular}[c]{@{}c@{}}\textbf{Holdout} \\  \textbf{Model(Sim)}\end{tabular}} 
    & \multicolumn{1}{c|}{\begin{tabular}[c]{@{}c@{}}\textbf{Holdout} \\  \textbf{(Live)}\end{tabular}} \\ 
    \hline
    	Partial     & 19.9mm          & 21.1mm          & 16.6mm          & 18.6mm          \\ \hline
    	Convex Hull & 13.9mm          & 16.1mm          & 14.1mm          & 10.5mm          \\ \hline
    	GPIS        & 17.1mm          & 16.0mm          & 21.3mm          & 20.8mm          \\ \hline
    	Depth CNN   & 12.1mm          & 13.7mm          & 12.4mm          & 22.9mm          \\ \hline
    	Ours        & \textbf{7.7mm} & \textbf{13.9mm} & \textbf{13.6mm} & \textbf{6.2mm} \\ \hline
    \end{tabular}
    
    \caption{\textbf{Pose error results} from simulated grasping experiments. This is the average L2 difference between planned and realized grasp pose averaged over the 3 finger tips and the palm of the hand, in millimeters. (Smaller is better) }
    \label{tab:sim_grasp_results} 
\end{subtable}&

\begin{subtable}[t]{.48\linewidth}

    \begin{tabular}{|c|c|c|c|c|c|c|}
    \hline
    \multicolumn{1}{|c|}{\begin{tabular}[c]{@{}c@{}}\textbf{Completion} \\  \textbf{Method}\end{tabular}} 
    & \multicolumn{1}{c|}{\begin{tabular}[c]{@{}c@{}}\textbf{Train} \\  \textbf{View(Sim)}\end{tabular}} 
    & \multicolumn{1}{c|}{\begin{tabular}[c]{@{}c@{}}\textbf{Holdout} \\  \textbf{View(Sim)}\end{tabular}} 
    & \multicolumn{1}{c|}{\begin{tabular}[c]{@{}c@{}}\textbf{Holdout} \\  \textbf{Model(Sim)}\end{tabular}} 
    & \multicolumn{1}{c|}{\begin{tabular}[c]{@{}c@{}}\textbf{Holdout} \\  \textbf{(Live)}\end{tabular}} \\ 
    \hline
    	Partial     & 8.19$^{\circ}$          & 6.71$^{\circ}$          & 8.78$^{\circ}$          & 7.67$^{\circ}$          \\ \hline
    	Convex Hull & 3.53$^{\circ}$          & 4.01$^{\circ}$          & 4.59$^{\circ}$          & 3.77$^{\circ}$          \\ \hline
    	GPIS        & 4.65$^{\circ}$          & 4.79$^{\circ}$          & 4.95$^{\circ}$          & 5.92$^{\circ}$          \\ \hline
    	Depth CNN   & 3.09$^{\circ}$          & 3.56$^{\circ}$          & 4.52$^{\circ}$          & 6.83$^{\circ}$          \\ \hline
    	Ours        & \textbf{2.48$^{\circ}$} & \textbf{3.41$^{\circ}$} & \textbf{4.95$^{\circ}$} & \textbf{2.43$^{\circ}$} \\ \hline
    \end{tabular}
    
    \caption{\textbf{Joint error results} from simulated grasping experiments. This is the mean L2 distance between planned and realized grasps in degrees averaged over the hand's 7 joints. Our method' smaller error demonstrates a more accurate geometry reconstruction. (Smaller is better)}
    \label{tab:l2_joint_error} 
\end{subtable}\\

    
    
    

\end{tabularx}
\end{table*}

\section{Completion of YCB/Grasp Dataset Objects}

We used the dataset from \cite{varley2017shapecompletion_iros} to create a new dataset consisting of half a million triplets of oriented voxel grids: depth, tactile, and ground truth. Depth voxels are marked as occupied if visible to the camera. Tactile voxels are marked occupied if tactile contact occurs within the voxel. Ground truth voxels are marked as occupied if the object intersects a given voxel, independent of perspective. The point clouds for the depth information were synthetically rendered in the Gazebo \cite{koenig2004design} simulator. This dataset consists of 608 meshes from both the Grasp \cite{kappler2015leveraging} and YCB \cite{calli2015ycb} datasets. 486 of these meshes were randomly selected and used for a training set and the remaining 122 meshes were kept for a holdout set. 

The synthetic tactile information was generated according to Algorithm \ref{alg:TactileDataGenerationUniform}. In order to generate tactile data, the voxelization of the ground truth high resolution mesh (vox\_gt) (Alg.\ref{alg:TactileDataGenerationUniform}:L\ref{line:alg1line1}) was aligned with the captured depth image (Alg.\ref{alg:TactileDataGenerationUniform}:L\ref{line:alg1line4}). 40 random $(x,y)$ points were sampled in order to generate synthetic tactile data (Alg.\ref{alg:TactileDataGenerationUniform}:L\ref{line:alg1line5}-\ref{line:alg1line6}).  For each of these points (Alg.\ref{alg:TactileDataGenerationUniform}:L\ref{line:alg1line7}), a ray was traced in the $-z$, direction and the first occupied voxel was stored as a tactile observation  (Alg.\ref{alg:TactileDataGenerationUniform}:L\ref{line:alg1line11}). Finally this set of tactile observations was converted back to a point cloud (Alg.\ref{alg:TactileDataGenerationUniform}:L\ref{line:alg1line13}).  

Two identical CNNs were trained where one CNN was provided only depth information (\textbf{Depth Only}) and a second was provided both tactile and depth information (\textbf{Tactile and Depth}).  During training, performance was evaluated on simulated views of meshes within the training data (\textit{Training Views}), novel simulated views of meshes in the training data (\textit{Holdout Views}), novel simulated views of meshes not in the training data (\textit{Holdout Meshes}), and real non-simulated views of 8 meshes from the YCB dataset (\textit{Holdout Live}). 

The \textit{Holdout Live} examples consist of depth information captured from a real Kinect and tactile information captured from a real Barrett Hand attached to a Staubli Arm.  We used depth filtering to mask out the background of the captured depth cloud. The object was fixed in place during the tactile data collection process. While collecting the tactile data, the arm was manually moved to place the end effector behind the object and 6 exploratory guarded motions were made where the fingers closed towards the object. Each finger stopped independently when contact was made with the object, as shown in Fig. \ref{fig:handobjectcontact}. 


Fig. \ref{fig:jaccard_improvement} demonstrates that the difference between the \textbf{Depth Only} CNN completion and the \textbf{Tactile and Depth} CNN completion becomes larger on more difficult completion problems. The performance of the \textbf{Depth Only} CNN nearly matches the performance of the \textbf{Tactile and Depth} CNN on the training views. Because these views are used during training, the network is capable of generating reasonable completions. Moving from \textit{Holdout Views} to \textit{Holdout Meshes} to \textit{Holdout Live}, the completion problems move further away from the examples experienced during training.  As the problems become harder, the \textbf{Tactile and Depth} network outperforms the \textbf{Depth Only} network by a greater margin, as it is able to utilize the sparse tactile information to differentiate between various possible completions. This trend shows that the network is able to make more use of the tactile information when the depth information alone is insufficient to generate a quality completion. We generated meshes from the output of the combined tactile and depth CNN using a marching cubes algorithm. We also preserve the density of the rich visual information and the coarse tactile information by utilizing the post-processing from \cite{varley2017shapecompletion_iros}.  


\begin{figure*}[t]
\vspace{2mm}

    \centering
    \includegraphics[width=\textwidth]{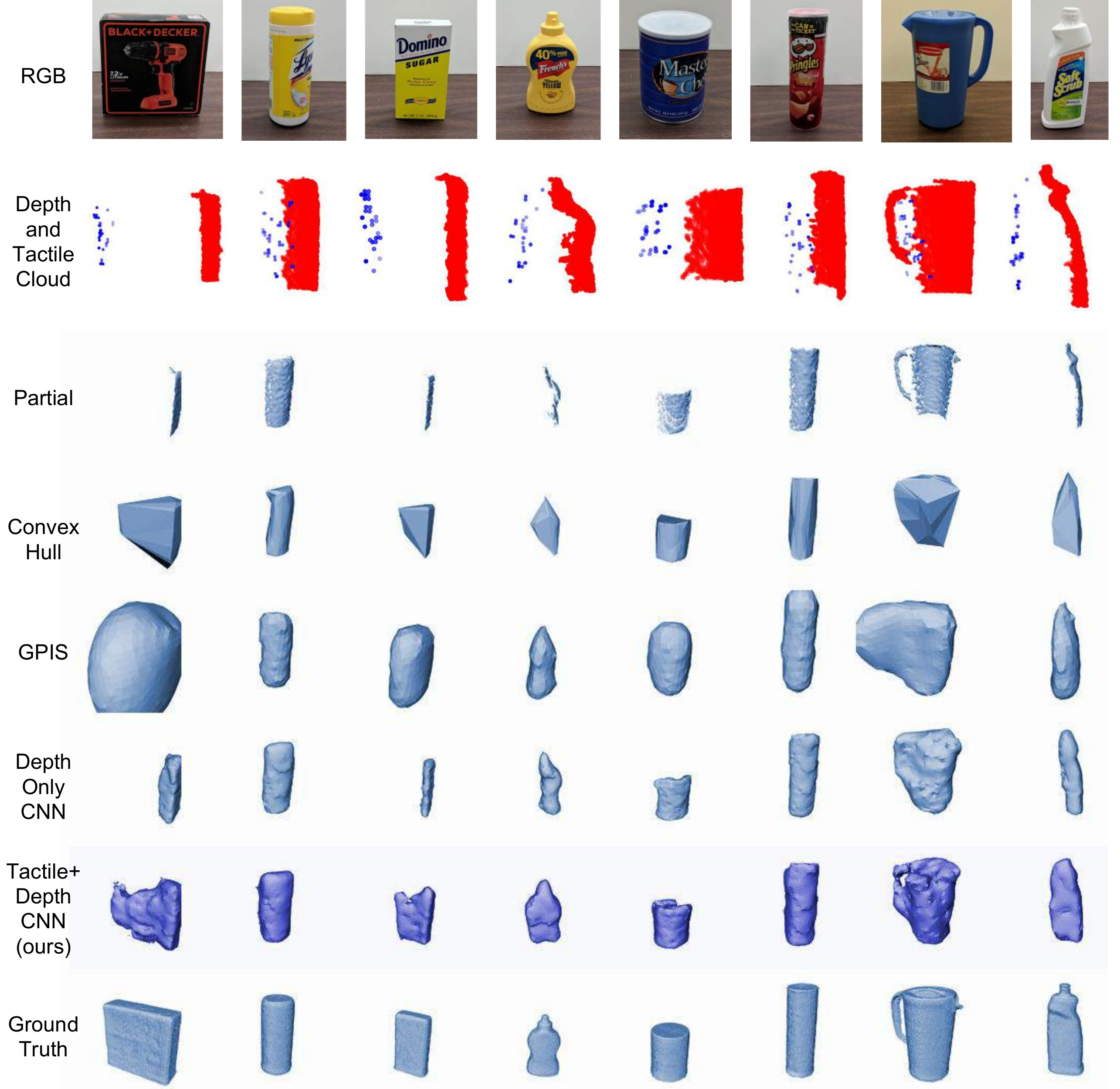}
\caption{The entire \textit{Holdout Live} dataset.  These completions were all created from data captured from a real Kinect and a real Barrett Hand attached to a Staubli Arm. The \textbf{Depth and Tactile Clouds} have the points captured from a Kinect in red and points captured from tactile data in blue.  Notice many of the \textbf{Depth Only} completions do not extend far enough back but instead look like other objects that were in the training data (ex: cell phone, banana). Our method outperforms the \textbf{Depth Only}, \textbf{Partial}, and \textbf{Convex Hull} methods in terms of Hausdorff distance and Jaccard similarity. Note that the \textbf{GPIS} completions form large and inaccurate completions for the Black and Decker box and the Rubbermaid Pitcher, whereas our method correctly bounds the end of the box and finds the handle of the pitcher. }
\label{fig:live_completions_full_dataset} 
\end{figure*} 

\setcounter{table}{4}
\begin{table}[t]
\vspace{2mm}

    \begin{tabular}{|c|c|c|c|c|c| }
    \hline
    \multicolumn{1}{|c|}{\begin{tabular}[c]{@{}c@{}}\textbf{Completion} \\  \textbf{Method}\end{tabular}} 
    & \multicolumn{1}{c|}{\begin{tabular}[c]{@{}c@{}}\textbf{Partial}\end{tabular}} 
    & \multicolumn{1}{c|}{\begin{tabular}[c]{@{}c@{}}\textbf{Convex} \\ \textbf{Hull}\end{tabular}} 
    & \multicolumn{1}{c|}{\begin{tabular}[c]{@{}c@{}}\textbf{GPIS}\end{tabular}} 
    & \multicolumn{1}{c|}{\begin{tabular}[c]{@{}c@{}}\textbf{Depth} \\ \textbf{CNN}\end{tabular}} 
    & \multicolumn{1}{c|}{\begin{tabular}[c]{@{}c@{}}\textbf{Ours}\end{tabular}} 
     \\ \hline
    
        Lift Success (\%) & 62.5\%& 62.5\%& 87.5\%& 75.0\%&\textbf{87.5\%}\\ \hline
        Joint Error ($^{\circ}$) & 6.37$^{\circ}$& 6.05$^{\circ}$& 10.61$^{\circ}$& 5.42$^{\circ}$& \textbf{4.67$^{\circ}$} \\ \hline
        Time (s) & 1.533s & \textbf{0.198s} & 45.536s & 3.308s & 3.391s\\ \hline
    \end{tabular}
    
    \caption{\textbf{Lift Success} is the percentage of successful lift executions. \textbf{Joint Error} is the average error per joint in degrees between the planned and executed grasp joint values. While GPIS and our method have the same lift success, our method is 1340\% faster and has 41\% of the joint error, making the process more reliable. (Smaller is better). \textbf{Average time to complete a mesh} using each completion method. While the convex hull completion method is fastest, ours has a superior tradeoff between speed and quality. (Smaller is better)}
    \label{tab:live_grasp_results} 
    
    \vspace{-2mm}

\end{table}

\section{Comparison to Other Completion Methods}
\label{sec:Completion_results}

\subsection{Alternative Visual-Tactile Completion Methods}
In this work we benchmarked our framework against the following general visual tactile completion methods.\\ 
\indent\textbf{Partial Completion}: The set of points captured from the Kinect is concatenated with the tactile data points. The combined cloud is run through marching cubes, and the resulting mesh is then smoothed using Meshlab's \cite{cignoni2008meshlab} implementation of Laplacian smoothing.  These completions are incredibly accurate where the object is directly observed but make no predictions in unobserved areas of the scene. \\
\indent\textbf{Convex Hull Completion}: The set of points captured from the Kinect is concatenated with the tactile data points.  The combined cloud is run through QHull to create a convex hull.  The hull is then run through Meshlab's implementation of Laplacian smoothing. These completions are reasonably accurate near observed regions. However, a convex hull will fill regions of unobserved space.\\ 
\indent\textbf{Gaussian Process Implicit Surface Completion (GPIS)}: Approximated depth cloud normals were calculated using PCL's KDTree normal estimation. Approximated tactile cloud normals were computed to point towards the camera origin. The depth point cloud was downsampled to size $M$ and appended to the tactile point cloud. We used a distance offset $d$ to add positive and negative observation points along the direction of the surface normal. We then sampled the Gaussian process using \cite{gerardo2014robust} with a $n^3$ voxel grid and a noise parameter $s$ to create meshes from the point cloud. We empirically determined the values of $M, s, n, d$ by sampling the Jaccard similarity of GPIS completions where $M=[200, 300, 400]$, $s=[0.001, 0.005]$, $n=[40, 64, 100]$, and $d=[0.005, 0.0005]$. We found $M=300$ to be a good tradeoff between speed and completion quality. Additionally we used $s=0.001$, $d=0.0005$, and $n=100$. 

In prior work \cite{varley2017shapecompletion_iros} the Depth Only CNN completion method was compared to both a RANSAC based approach \cite{papazov2010efficient} and a mirroring approach \cite{bohg2011mind}. These approaches make assumptions about the visibility of observed points and do not work with data from tactile contacts that occur in unobserved regions of the workspace. 

\subsection{Geometric Comparison Metrics}
The Jaccard similarity was used to compare $40^3$ CNN outputs with the ground truth. We also used this metric to compare the final resulting meshes from several completion strategies. The completed meshes were voxelized at $80^3$ and compared with the ground truth mesh. The results are shown in Table \subref{tab:Jaccard}. Our proposed method results in higher similarity to the ground truth meshes than do all other described approaches. 

The Hausdorff distance metric computes the average distance from the surface of one mesh to the surface of another. A symmetric Hausdorff distance was computed with Meshlab's Hausdorff distance filter in both directions. Table \subref{tab:Hausdorff} shows the mean values of the symmetric Hausdorff distance for each completion method. In this metric, our tactile and depth CNN mesh completions are significantly closer to the ground truth compared to the other approaches' completions.  


Both the partial and Gaussian process completion methods are accurate when close to the observed points but fail to approximate geometry in occluded regions. We found that in our training, the Gaussian Process completion method would often create a large and unruly object if the observed points were only a small portion of the entire object or if no tactile points were observed in simulation. Using a neural network has the added benefit of abstracting object geometries, whereas the alternative completion methods fail to approximate the geometry of objects which do not have points bounding their geometry. 

\subsection{Grasp Comparison in Simulation}

In order to evaluate our framework's ability to enable grasp planning, the system was tested in simulation using the same set of completions. The use of simulation allowed for the quick planning and evaluation of  7900 grasps. GraspIt! was used to plan grasps on all of the completions of the objects by uniformly sampling different approach directions. These grasps were then executed not on the completed object but on the ground truth meshes in GraspIt!. In order to simulate a real-world grasp execution, the completion was removed from GraspIt! and the ground truth object was inserted in its place. Then the hand was placed 20 cm away from the ground truth object along the approach direction of the grasp. The spread angle of the fingers was set, and the hand was moved along the approach direction of the planned grasp either until contact was made or a maximum approach distance was traveled. At this point, the fingers closed to the planned joint values. Then each finger continued to close until either contact was made with the object or the joint limits were reached. 

Table \subref{tab:sim_grasp_results} shows the average difference between the planned and realized Cartesian finger tip and palm poses, while Table \subref{tab:l2_joint_error} shows the difference in pose of the end effector between the planned and realized grasps averaged over the 7 joints of the hand. Using our method, the end effector ended up closer to its intended location in terms of both joint space and the palm's Cartesian position versus other completion methods' grasps.

\subsection{Live Grasping Results}

To further test our network's efficacy, the grasps were planned and executed on the Holdout Live views using a Staubli arm with a Barrett Hand. The grasps were planned using meshes from the different completion methods described above. For each of the 8 objects, we ran the arm once using each completion method. The results are shown in Fig. \ref{fig:live_completions_full_dataset} and Table \ref{tab:live_grasp_results}. Our method enabled an improvement over the other visual-tactile shape completion methods in terms of grasp success rate and resulted in executed grasps closer to the planned grasps, as shown by the lower average joint error (and much faster than GPIS). 
\section{Conclusion}

\label{sec:conclusion}

Our method provides an open source novel visual-tactile completion method which outperforms other general visual-tactile completion methods in completion accuracy, time of execution, and grasp posture utilizing a dataset which is representative of household and tabletop objects. We  demonstrated  that  even  small  amounts  of  additional tactile information can be incredibly helpful in reasoning about object geometry. This CNN uses both dense depth information and sparse tactile information to fill in occluded regions of an object. Experimental results verified that utilizing both vision and tactile was superior to using depth alone. In the future we hope to relax the fixed object assumption by using tactile sensors developed in our lab that allow contact without motion. We are also interested in a more general exploration algorithm of the unseen part using tactile.

\clearpage

\bibliographystyle{IEEEtran}
\bibliography{bib/references}

\end{document}